\documentclass[11pt]{article}

\usepackage[preprint]{acl}

\usepackage{times}
\usepackage{latexsym}

\usepackage[T1]{fontenc}

\usepackage[utf8]{inputenc}

\usepackage{microtype}

\usepackage{inconsolata}

\usepackage{graphicx}

\usepackage{amsmath}
\usepackage{booktabs}
\usepackage{enumitem}
\usepackage{booktabs, rotating, multirow}
\usepackage{pifont}

\newcommand{\cmark}{\ding{51}} \newcommand{\xmark}{\ding{55}}

\title{Where Hallucinations Live: A Cross-Architecture Circuit in VQ-Tokenized Vision-Language Models}

\author{
  \textbf{Shamanthak Hegde}\textsuperscript{1}\thanks{Corresponding author: \texttt{shamanthak@asu.edu}}
  \textbf{Xiangrui Liu}\textsuperscript{1}
  \textbf{Maitreya Patel}\textsuperscript{2}\thanks{Partial work done while at Arizona State University.}
  \textbf{Yezhou Yang}\textsuperscript{1}
\\
\vspace{0.5em}
  \textsuperscript{1}Arizona State University \qquad \textsuperscript{2}Adobe \\
  \url{https://shamanthak-hegde.github.io/where-hallucinations-live}
}

\begin{document}
\maketitle

\begin{abstract}

Unified vision-language models (VLMs) that tokenize images through a vector-quantized (VQ) codebook routinely hallucinate objects on grounded yes/no benchmarks, yet existing decoding-time fixes treat this as generic miscalibration without an architectural account. Using activation patching across twenty-five models spanning eight LLM families, we identify an early-layer ($L_0$) attention routing circuit shared across VQ-tokenized VLMs and propose a three-gate diagnostic that distinguishes the models carrying it from those that do not. The diagnostic isolates ten positive models (five natural unified-VQ VLMs across three LLM families and five induced variants) and rejects the remaining fifteen. A single-variable architectural swap (LLaVA-1.6 CLIP+MLP $\rightarrow$ VQ+Linear) installs the circuit, while a matched-compute MLP control on identical data does not, isolating vector quantization as the source of the pathological signal; the routing pathway that carries it is one that the backbone already provides. Against tuned VCD and DoLA baselines, tuned DoLA wins on binary calibration, but \textbf{only $L_0$ ablation reduces object hallucination in open-ended generation} (CHAIR$_i$ reduces by $31\,\%$ relatively, whereas tuned DoLA and VCD leave it unchanged or worsen it). These results recast object hallucination in unified VQ VLMs as a property of architecture and pretraining, and yield a targeted intervention that mechanism-agnostic decoding cannot replicate. 

\end{abstract}
\section{Introduction}
\label{sec:intro}

Unified vision-language models that share a single transformer between image and text tokens have become a dominant architecture~\citep{vilau2024, chameleon2024, liquid2025, chern2024anole, liu2024lumina-mgpt, janus2024, showo2024, emu3, jin2024unified}. Most of them tokenize images through a vector-quantized (VQ) codebook so visual tokens occupy the same vocabulary as text. While elegant, this design suffers from a persistent failure mode on grounded yes/no benchmarks such as POPE~\citep{pope2023} and AMBER~\citep{amber2023}: VQ-tokenized models exhibit extreme and architecturally consistent biases. Several, including VILA-U~\citep{vilau2024} and Liquid~\citep{liquid2025}, answer "\textsc{yes}" on roughly $78-81\%$ of records regardless of image content, while others collapse to the opposite extreme: Chameleon~\citep{chameleon2024} answers "\textsc{yes}" on only $22\%$ of records, and Lumina-mGPT~\citep{liu2024lumina-mgpt} on $0.07\%$. Continuous-projector VLMs such as LLaVA-1.6~\citep{liu2024llavanext} and Qwen2.5-VL~\citep{Qwen2.5-VL} show none of these extreme patterns. The signature is too consistent within the VQ family to be coincidental, and too at odds with the continuous family to be a generic VLM defect, pointing to a shared architectural cause.

Existing interventions for VLM hallucination such as VCD~\citep{vcd2024}, DoLA~\citep{dola2024}, ITI~\citep{iti2023}, VTI~\citep{liu2024reducing}, operate on output logits without an architectural account. These methods are mechanism-agnostic: each targets a downstream symptom (per-token calibration, layer-contrast logits, residual-direction steering) without explaining \emph{why} a particular family of models hallucinates in the first place. They are also evaluated almost exclusively on binary benchmarks, where a single-token forced-choice makes logit-level fixes maximally effective; whether they actually reduce object hallucination in open-ended generation is a separate question. Recent mechanistic studies of individual VLMs have begun to address the \emph{why} \citep{transformer_circuits2022, rome2022, iioi2023,liu2025investigating}, but typically as single-model case studies. Whether a \emph{shared} circuit, a specific attention pathway through which information flows to causally produce a behavior, underlies hallucination \emph{across} VQ-tokenized architectures, whether such a circuit can be detected before training a model, and whether targeting it yields interventions that beat mechanism-agnostic baselines on generative metrics remain unanswered.

We answer these questions affirmatively. Using activation patching~\cite{rome2022} across twenty-five models spanning eight LLM families, we identify an early-layer ($L_0$) attention routing circuit present across five unified-VQ VLMs spanning three LLM families. The circuit is absent in continuous-projector models and is discriminable from other unified-VQ VLMs that fail the diagnostic for mechanistically distinct reasons. It can be constructed from a healthy backbone by a single-variable swap, replacing LLaVA-1.6's CLIP+MLP projector with a VQ+Linear projector; a matched-compute MLP variant on identical data and budget fails to install it, isolating the VQ bottleneck as the cause.


To formalize the cross-architecture pattern, we propose a three-gate diagnostic combining residual-stream divergence, attention-routing concentration at $L_0$, and codebook off-manifold rate, with thresholds tied to architectural properties measured before behavioral outcomes (Section~\ref{sec:diagnostic}). The mechanism also yields a targeted intervention. Sweeping VCD and DoLA over $36$ configurations on VILA-U reveals a striking dissociation: tuned DoLA wins on binary POPE/AMBER, but \textbf{only $L_0$ ablation reduces object hallucination in open-ended generation} (CHAIR$_i$ $10.3\,\%\!\rightarrow\!7.1\,\%$). Logit-level fixes calibrate the yes/no decision but do not disrupt the $L_0$ circuit that injects hallucinated objects during generation. Our main contributions are summarized as follows:

\begin{itemize}[leftmargin=*,itemsep=1pt,topsep=2pt]
\item A cross-architecture $L_0$ routing circuit present in five natural unified-VQ VLMs across three LLM families, absent in continuous-projector models, and structurally validated on a held-out model not used to calibrate the diagnostic.

\item Causal identification of the VQ bottleneck via single-variable construction with a matched-compute MLP control, and a training-time asymmetry showing the circuit is encoded in LLM backbone weights.

\item A three-gate diagnostic chain with an architecturally motivated (not threshold-fit) two-tier criterion that correctly discriminates twenty-five models with mechanistically distinct positive and negative signatures.

\item A mechanism-targeted intervention that loses to tuned VCD/DoLA on binary calibration but uniquely reduces object hallucination in open-ended generation.
\end{itemize}

\section{Background and Related Work}
\label{sec:related}
\paragraph{Vector quantization in VLMs.}
VQ tokenizers~\citep{vqvae2017, vqgan2021, rqvae2022, fsq2024, lfq2024} map each patch of a continuous visual encoder's output to its nearest entry in a learned codebook of size $K$, so visual content becomes discrete tokens sharing a vocabulary with text. Two failure modes recur: codebook collapse, where only a small fraction of the $K$ codes are used in practice (Section~\ref{sec:c2}); and snapping artifacts, where a small input perturbation flips the output to an unrelated code with no smooth gradient between them, a property that becomes load-bearing for the hallucination mechanism we develop.

\paragraph{Hallucination correction in VLMs.}
Decoding-time methods correct hallucination through logit-level interventions at inference, without modifying weights: VCD~\citep{vcd2024}, DoLA~\citep{dola2024}, ITI~\citep{iti2023}, and VTI~\citep{liu2024reducing}. All are architecture-agnostic and offer no account of \emph{why} a given model hallucinates. A separate line of mechanistic-interpretability work addresses the \emph{why}, but as single-model case studies. We extend mechanistic analysis to cross-architecture generality and compare against fully tuned versions of these baselines.
\paragraph{Activation patching and routing circuits.}
Activation patching~\citep{rome2022, iioi2023, transformer_circuits2022} localizes where behavior is causally encoded: one runs the model on a clean input and a corrupted variant, then re-runs the corrupted input while patching in clean activations one site at a time. Our diagnostic (Section~\ref{sec:diagnostic}) builds on this procedure.
\section{Method}
\label{sec:method}

\begin{figure*}
    \centering
    \includegraphics[width=0.95\linewidth]{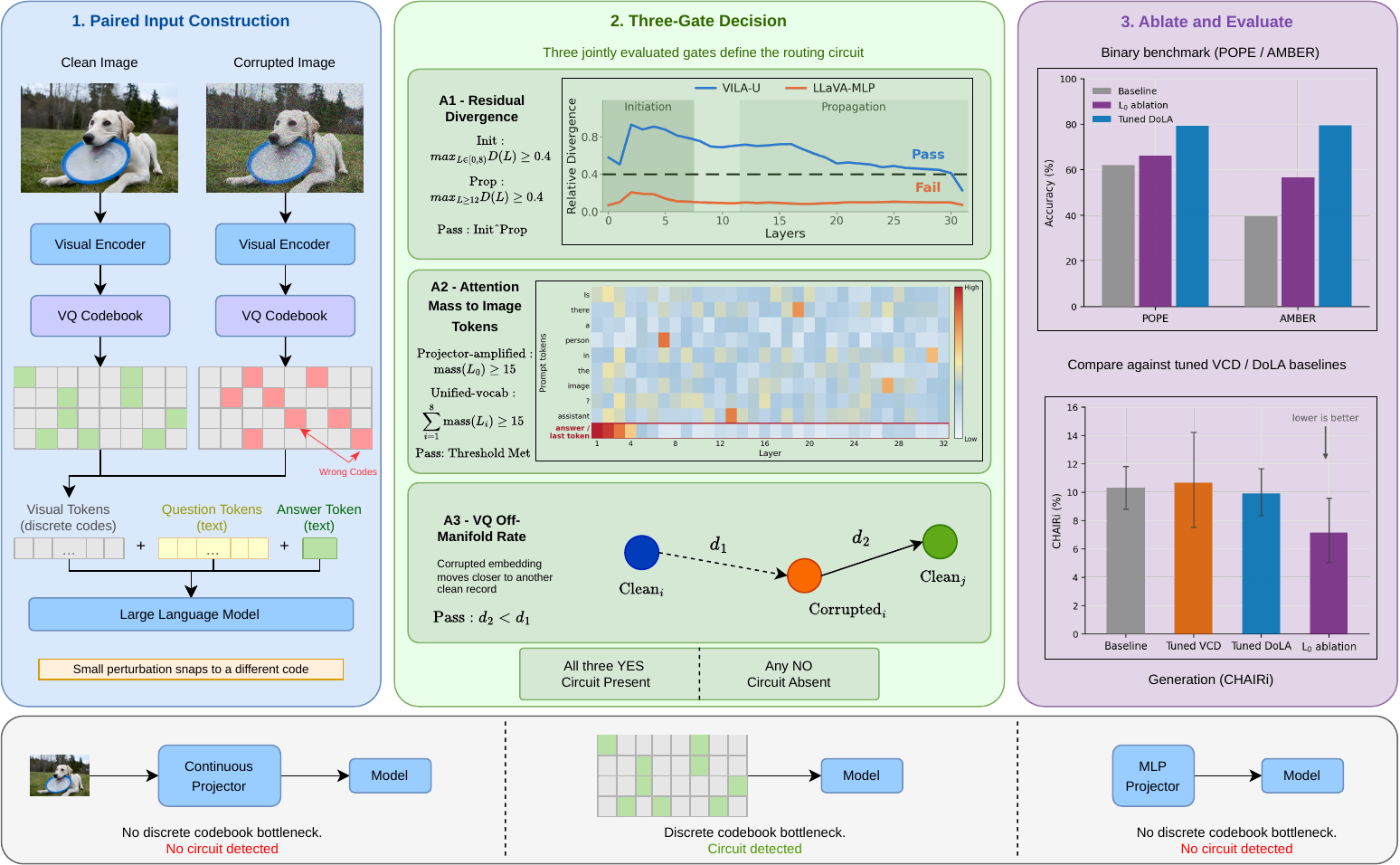}
    \caption{Methodology overview. Clean and corrupted image pairs are constructed and passed through the transformer; residual divergence and early-layer attention mass are measured to detect a routing circuit via a three-gate diagnostic. Models passing all three gates have a VQ-installed $L_0$ circuit confirmed by a single-variable causal control: swapping only the projector type from MLP to VQ installs the circuit; a matched-compute MLP control does not. Ablating $L_0$ reduces object hallucination in open-ended generation (CHAIR$_i$ drops 31\% relative) while tuned DoLA achieves better binary calibration.}
    \label{fig:method}
\end{figure*}
 
With reference to Figure~\ref{fig:method}, our method has three components, each motivated by a question the preceding component leaves open: (i) a mechanistic diagnostic that identifies whether a given VLM has the candidate routing circuit; (ii) a constructive induction procedure with a matched-compute control that tests whether the circuit is causally tied to the VQ bottleneck, rather than to the incidental features of the models that happen to exhibit it; and (iii) a sanity criterion that distinguishes mechanism-targeted reductions in hallucination from numerical improvements driven by the model failing to respond at all. 

\subsection{Mechanistic diagnostic: a three-gate chain}
\label{sec:diagnostic}

We instantiate the activation-patching procedure of Section~\ref{sec:related} with two corruption types: Gaussian noise on visual embeddings and image-swap corruption, applied across a fixed set of 500 paired records (300 NaturalBench~\citep{naturalbench2024} image-swap pairs and 200 POPE~\citep{pope2023} Gaussian-noise pairs). For each record, we run the model on the clean visual embedding $v$ and, depending on corruption type, either a noise version $v + \sigma \cdot \epsilon$ where $\epsilon \!\sim\! \mathcal{N}(0, I)$ or a swapped-image embedding, caching activations at every layer; we then re-run the corrupted input while patching in the clean activation at one (token-category, layer) site at a time, with categories visual, question, and answer grouping image-embedding, prompt, and response positions. The restoration score $R = (\mathrm{logit}_{\text{patch}} - \mathrm{logit}_{\text{corrupt}}) /(\mathrm{logit}_{\text{clean}} - \mathrm{logit}_{\text{corrupt}})$ is read out on the answer token: $R=0$, $1$, $>\!1$ correspond to no effect, full restoration, and overshoot.

\paragraph{Per-model noise calibration.}
Visual-embedding norms vary by over an order of magnitude across our cohort, so a fixed $\sigma$ would over-perturb some models and under-perturb others. For each model, we sweep $\sigma$ over a logarithmic grid and select the value whose mean per-token cosine similarity between clean and noisy embeddings, averaged over a subset of POPE probe records, is closest to $0.5$; this operates purely on input embeddings, independent of model predictions and labels. Accordingly, $\sigma$ sorts our cohort into three architectural classes (\emph{$\sigma$-tiers}): \emph{projector-amplified VQ} (high-norm; e.g., VILA-U, Liquid), \emph{continuous-projector} (e.g., LLaVA-1.6, Qwen2.5-VL), and \emph{unified-vocab VQ} (low-norm; e.g., the Chameleon family, where visual tokens enter the LM's token-embedding table at text-token scale).

\paragraph{Robustness to corruption type.}
The early-layer divergence that $A1$ measures is not specific to Gaussian embedding noise: token dropout reproduces the early peak in both tiers, while permuting the visual-token embeddings produces near-zero divergence despite carrying the largest perturbation magnitude of the corruptions tested, indicating that the routing pools visual-token content independently of order (Appendix~\ref{app:corruption_type}).

\paragraph{The three gates.}
A model passes the diagnostic if it satisfies three gates jointly. 
\textbf{Gate $A1$: early routing event with downstream propagation.} We measure the per-layer relative divergence between clean and corrupted residual streams ($\ell_2$ distance between hidden states, normalized by the clean hidden-state norm). The candidate circuit predicts two things: a routing event concentrated in the early layers, and persistence of the resulting signal rather than its dissipation. $A1$ therefore has two parts. $A1$-initiation requires the relative divergence to reach $\geq\!0.4$ within the early layers $[0, 8)$, the first quarter of the stack. $A1$-propagation requires it to reach $\geq\!0.4$ somewhere in the layers $\ell \geq 12$, so that the signal is still carried in the second half of the network. The two measurements are taken on disjoint ranges separated by a four-layer buffer, so neither can be satisfied by the decaying shoulder of the other; the gate tests sustained divergence rather than two separate events, and in several passing models, the elevated divergence is a single continuous ramp or plateau whose onset and persistence the two measurements probe. A model passes $A1$ only if both hold. A perturbation that produces no early event (initiation fails) reflects a model that does not route visual information through its early layers; one that rises early and then decays below threshold (propagation fails) reflects a routing event that is causally inert. Neither bound is tuned to the cohort partition: the verdict partition is invariant across initiation windows $[0, 4)$ and $[0, 8)$, while at $\ell \geq 8$ the propagation maximum for two positive models falls on the decaying shoulder of the initiation peak, and at $\ell \geq 16$ a positive model drops below threshold. Section~\ref{sec:c5} shows that both failure modes are populated by mechanistically distinct negatives. 
\textbf{Gate $A2$: early-layer attention concentration.} We measure the fraction of attention mass at layer $0$ that the prompt-last token directs to visual tokens, summed over heads. Because the three $\sigma$-tiers operate at different embedding-norm regimes, $A2$ applies a two-tier criterion: $L_0$ mass $\geq\!15$ for projector-amplified models; window total over $[0, 8)$ of $\geq\!15$ for unified-vocab models. The two-tier structure is inherited from the $\sigma$-tier classification (defined from embedding norms before any behavioral outcome is measured) rather than fit to cohort labels. Continuous-projector and decoupled models, which have no early-routing pathway, are measured at the layer-0 sink like projector-amplified models and fall well below the threshold. 
\textbf{Gate $A3$: codebook off-manifold rate.} For each record, we check whether the noised visual embedding is closer to a different record's nearest codebook entry than to its own. $A3$ requires this rate to be $\geq\!80\,\%$, capturing the snapping artifact at the cohort scale. A model that cannot be measured at a gate does not pass it, except where the gate is explicitly waived. Where the quantizer exists but uses a fixed lattice, as in the FSQ variants, $A3$ is waived; where the understanding pathway is not quantized at all, the premise the gate tests is absent, and the model is rejected at this gate.

\paragraph{Why a chain.}
The gates are applied as a conjunctive chain because, individually, they admit confounds. $A2$ alone can be passed by a model that routes visual tokens through $L_0$ without driving a pathological signal along that route: the matched-compute MLP control of Section~\ref{sec:c2} passes $A2$ in isolation, a healthy backbone already routes early visual attention through $L_0$, yet fails $A1$ and shows a null behavioral $L_0$ effect. Two natural negatives make the same point: UniTok and Emu3 both clear the $A2$ mass threshold yet fail $A1$, UniTok for lack of an early event and Emu3 for lack of propagation. Qwen-VQ makes the converse point, passing $A1$ and $A3$ and failing only the mass screen. Only the conjunction of a routing event ($A2$), a divergence signal that initiates early and propagates downstream ($A1$), and codebook off-manifold behavior ($A3$) recovers the correct verdict.
 
\subsection{Constructive induction procedure with matched-compute control}
\label{sec:induction-protocol}

The cross-architecture pattern across natural unified-VQ models admits a confounding explanation: those models differ from the continuous-projector models in many ways besides the tokenizer. To isolate vector quantization, we run a single-variable contrast. Starting from a healthy continuous-projector backbone (LLaVA-1.6 with Vicuna-7B and CLIP), we freeze the LM and the CLIP encoder and train \emph{only} the projector under two configurations that differ in exactly one component. The first replaces the continuous CLIP+MLP projector with a VQ+Linear projector (a linear map with codebook quantization at capacity $K$). The second replaces it with a 2-layer GeLU MLP, same as LLaVA's original projector. Both configurations are trained for $2000$ steps on the same $50$k CC3M~\citep{cc3m} caption subset under identical hyperparameters (Appendix~\ref{app:hparams}); the only difference is whether the projector output is quantized through a codebook before reaching the LM.

If a brief projector-only training were itself sufficient to install the behavioral circuit, both configurations would do so. If the VQ bottleneck is the causally responsible component, only the VQ configuration should, and we treat this as the central causal test in Section~\ref{sec:c2}. Two additional sweeps probe how the circuit depends on codebook structure. We vary capacity $K\!\in\!\{1024, 4096, 16384, 65536\}$ to test whether $K$ alone predicts the circuit's \emph{polarity}, whether the model is biased toward \textsc{yes} or \textsc{no} answers under the resulting circuit. We also train two finite-scalar-quantization FSQ \citep{fsq2024} variants with different per-dimension level configurations (Appendix~\ref{app:hparams}). Because FSQ cannot exhibit codebook collapse by construction, the FSQ configuration tests whether the circuit forms in the absence of collapse. A second configuration was trained to distinguish a regime effect from a single-configuration artifact, but it fails $A1$ at initiation, so the comparison is between one circuit-installing configuration and one that is not.


\begin{table*}[t]
\centering
\caption{Five natural unified-VQ VLMs spanning three LLM families that pass the three-gate diagnostic. For $A1$, \textit{init} is the maximum relative divergence over layers $[0,8)$ and \textit{prop} the maximum over $\ell \ge 12$; both must reach 0.4. $A2$ is scored on $L_0$ mass for projector-amplified models (P) and the $[0,8)$ window total for unified-vocabulary models (U); both apply a $\ge 15$ threshold (Section~\ref{sec:diagnostic}). Behavioral 95\% CIs from 10,000 paired-record bootstrap resamples. $^\dagger$Lumina-mGPT is held out: baseline yes-rate at the behavioral floor.}
\label{tab:passspecs}
\resizebox{\textwidth}{!}{%
\begin{tabular}{llccccrr}
\toprule
& & \multicolumn{2}{c}{Gate $A1$} & & & \multicolumn{2}{c}{POPE yes-rate (\%)}\\
\cmidrule(lr){3-4}\cmidrule(lr){7-8}
Model & Backbone & init $[0,8)$ & prop $\ell\!\ge\!12$ & $A2$ (tier) & $A3$ & base$\rightarrow$L0 & $\Delta$ [95\% CI]\\
\midrule
VILA-U-7B            & Vicuna    & 0.932 ($L_2$) & 0.724 ($L_{16}$)  & 17.5 (P)  & 100 & 81.0$\rightarrow$72.6 & $-8.4$ [$-9.7$, $-7.2$]\\
Liquid-7B         & Gemma     & 0.768 ($L_1$) & 0.819 ($L_{24}$)  & 19.2 (U)  & 100 & 78.2$\rightarrow$\;0.0 & $-78.2$ [$-79.7$, $-76.7$]\\
Chameleon-7B         & Chameleon & 0.635 ($L_7$) & 7.267 ($L_{31}$)  & 22.0 (U)  & 100 & 22.3$\rightarrow$\;9.8 & $-12.5$ [$-14.3$, $-10.7$]\\
Anole-7B             & Chameleon & 2.009 ($L_1$) & 34.149 ($L_{31}$) & 111.2 (U) & 100 & 75.3$\rightarrow$\;2.0 & $-73.3$ [$-74.9$, $-71.7$]\\
Lumina-mGPT-7B$^\dagger$ & Chameleon & 2.807 ($L_6$) & 0.623 ($L_{20}$) & 175.0 (U) & 100 & 0.07$\rightarrow$\;0.0 & $-0.07$ [$-0.17$, 0.00]\\
\bottomrule
\end{tabular}
}
\end{table*}
 
\subsection{Interventions and a sanity criterion}
\label{sec:interventions}

\paragraph{Interventions.}
We evaluate three mechanism-targeted interventions on VILA-U, each implemented as a forward-pass hook that modifies activations during inference without retraining: $L_0$ ablation, $L_1$ ablation, and VTI. $L_\ell$ ablation zeroes the self-attention output at layer-$\ell$; we report $L_0$ (the candidate circuit) and $L_1$ (the downstream writer in models such as Chameleon, Section~\ref{sec:c6}). A separate window variant generalizes $L_\ell$ ablation to a contiguous range $[L_s, L_e)$ and is used only as a structural probe for Show-o and SEED-LLaMA, whose divergence has no single localized event to target, so no $L_0$ ablation is reported for them (Table~\ref{tab:failspecs}). \emph{VTI}~\citep{liu2024reducing}, described in Section~\ref{sec:related}, computes its steering direction from mean $\texttt{clean}-\texttt{corrupt}$ residual differences over demonstration records; we add it at every decoder block with $\alpha = 0.005$. VTI is prior work rather than a contribution of ours; we include it as a mechanism-adjacent comparison because its steering direction is built from the same $\texttt{clean}-\texttt{corrupt}$ residual differences the diagnostic uses.

\paragraph{A sanity criterion for hallucination reduction.} 
An intervention that improves a binary accuracy number is not automatically a hallucination reduction. Liquid's $L_0$ ablation, for instance, drives POPE yes-rate from $78\,\%$ to $0\,\%$ while producing all empty open-ended generations, the score improves because the model has stopped responding to the input, not because it has become better grounded. To rule out such cases, for each POPE record, we compute the logit gap $g = \mathrm{logit}_{\textsc{yes}} - \mathrm{logit}_{\textsc{no}}$ and apply a \emph{question-conditional} sanity criterion, computed separately on the yes-GT and no-GT record subsets, so a collapse affecting either alone is still detected. An intervention is \emph{preserved} only if both (a) the ratio of baseline to intervention within-class standard deviation of $g$ remains below $3$ (the model still varies across inputs), and (b) the across-class mean difference $\overline{g}_{\text{yes-GT}} - \overline{g}_{\text{no-GT}}$ remains $>\!0.5$ logits (the model still discriminates the two classes on average). An intervention failing either is a \emph{degenerate emitter}: it demonstrates that the targeted circuit is present (Section~\ref{sec:c1}) but not as a practical hallucination reduction. 

\subsection{Experimental setup}
\label{sec:setup}
We summarize the setup here; full details are in Appendix~\ref{app:hparams}. We evaluate on POPE, AMBER, and CHAIR~\citep{chair} ($500$ images). For activation patching, we use $500$ paired records ($300$ NaturalBench \texttt{image\_swap} + $200$ POPE \texttt{gaussian\_noise}), each pair consisting of one \textsc{yes} and one \textsc{no} record, which is what validates the paired-bootstrap CIs. A unified hook framework provides a consistent visual\,/\,question\,/\,answer\,/\,other token-category mapping across all models. All interval estimates reported in Appendix~\ref{app:bootstrap} use $10{,}000$ bootstrap resamples, resampling paired records on the binary benchmarks and images on CHAIR.

The studied models span natural VQ-VLMs (VILA-U~\citep{vilau2024}, Chameleon-7B~\citep{chameleon2024}, Liquid~\citep{liquid2025}, Anole~\citep{chern2024anole}, Lumina-mGPT~\citep{liu2024lumina-mgpt}, UniTok~\citep{unitok}, Show-o~\citep{showo2024}, SEED-LLaMA~\citep{ge2023making}, Emu3~\citep{emu3}, Janus-Pro~\citep{janus2024}), continuous-projector controls (LLaVA-1.6~\citep{liu2024llavanext}, VILA~\citep{lin2024vila}, Qwen2.5-VL~\citep{Qwen2.5-VL}, HaploOmni~\citep{xiao2025haploomni}, LaVIT~\citep{jin2024unified}, GILL~\citep{gill}), and induced variants (LLaVA-VQ at $K\!\in\!\{1024, 4096, 16384, 65536\}$, two LLaVA-VQ-FSQ configurations, Qwen-VQ, and a matched-compute LLaVA-MLP variant). Detailed gate values for all models appear in Appendix~\ref{app:cohort}
\section{Results}
\label{sec:results}

\begin{figure*}
    \centering
    \includegraphics[width=0.95\linewidth]{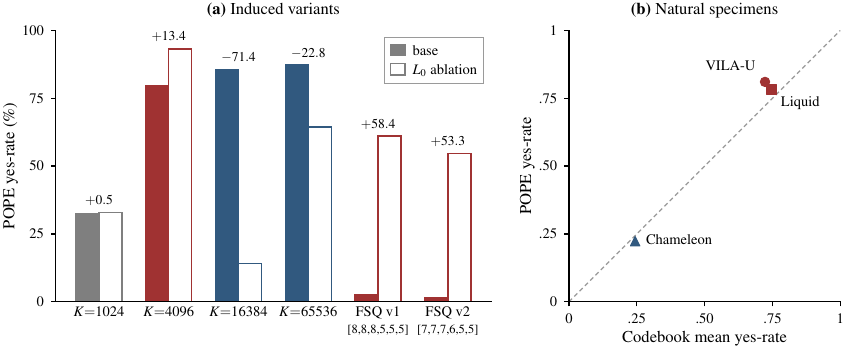}
    \caption{Codebook structure governs circuit polarity. (a) Induced LLaVA-VQ: capacity sweep and FSQ configurations, of which v1 installs the circuit and v2 does not. (b) Natural unified-VQ models: codebook mean yes-rate against behavioral baseline POPE yes-rate.}
    \label{fig:polarity}
\end{figure*}

\subsection{A shared early-layer routing circuit across VQ-tokenized VLMs}
\label{sec:c1}

The diagnostic chain identifies a shared $L_0$ attention routing circuit across five natural unified-VQ VLMs spanning three LLM families (Table~\ref{tab:passspecs}): Vicuna-7B in VILA-U, Gemma-7B~\citep{team2024gemma} in Liquid, and the Chameleon family in Chameleon-7B, Anole, and a held-out Lumina-mGPT. All five pass $A1$ (both halves), $A2$ under the two-tier criterion, and $A3$. Four exhibit large behavioral changes under $L_0$ ablation; Anole is particularly informative as an image-generation fine-tune of Chameleon-7B, with a gate-$A2$ window total five times larger than its base model, the circuit strengthens through generation fine-tuning, consistent with the broader claim that VQ-based training installs it. The layer-0 routing is a distributed pattern rather than a few-head circuit. Ranking the 32 heads at the routing layer by visual-to-prompt mass, thirteen heads are required to account for half of the total mass and twenty-two for eighty percent, with near-even per-head contributions (the strongest head carries 0.80 and the twenty-second still 0.63). No small subset dominates, which is why a full early-window knockout rather than a single-head ablation is required to disrupt the route (Section~\ref{sec:c6}), and why UniTok's diffuse layer-0 mass (Table~\ref{tab:a2-alternatives}) is disqualifying only in conjunction with its misaligned divergence event. ``Circuit'' here therefore denotes a coherent layer-level routing channel, not a sparse set of privileged heads.

Lumina-mGPT is held out: not in the threshold-calibration cohort. It passes all three gates with the strongest $A2$ signal observed in any model (Table~\ref{tab:passspecs}), yet its behavioral change under $L_0$ ablation is null, its baseline yes-rate is $0.07\,\%$, leaving no room to drop. This structural--behavioral dissociation is itself informative: the diagnostic detects a structural property of the architecture independent of whether the language prior happens to manifest it behaviorally, and the held-out model confirms the thresholds were not over-fit to the calibration cohort. Under the conjunctive definition of $A1$, Lumina-mGPT's binding value is its propagation maximum ($0.623$ at $L_{20}$) rather than its initiation peak, and that propagation value is the weakest among the natural models.

\paragraph{Polarity tracks codebook bias.}
The five models fall into three polarity regimes that align with their codebook's training distribution. VILA-U and Liquid have universally \textsc{yes}-biased codebooks and \textsc{yes}-promoting circuits; Chameleon's codebook is universally \textsc{no}-biased and its baseline yes-rate is $22\,\%$. The induced FSQ model of Section~\ref{sec:c2}, whose tokenizer cannot collapse by construction, produces a \textsc{yes}-suppressing circuit; a second FSQ configuration reproduces the polarity inversion behaviorally but does not install a gate-detectable early event. We frame this as associational rather than causal: Chameleon's $L_0$ still mildly promotes \textsc{yes} despite the codebook's \textsc{no}-bias (its overall \textsc{no} behavior comes from a downstream $L_1$ \textsc{no}-writer, Section~\ref{sec:c6}), and the LLaVA-VQ $K$-sweep is non-monotonic (Section~\ref{sec:c2}).



\begin{table*}[t]
\centering
\caption{The matched-compute contrast under identical training parameters on POPE. The VQ projector installs the $L_0$ routing circuit; the matched-compute MLP projector does not, failing both halves of $A1$, passing $A2$.}
\label{tab:matched_compute}
\begin{tabular}{lcc}
\toprule
\textbf{Metric} & \textbf{VQ+Linear} ($K\!=\!16384$) & \textbf{MLP} (matched-compute) \\
\midrule
$A1$ init $[0,8)$                & 1.276 \cmark  & 0.206 \xmark \\
$A1$ prop $\ell\!\ge\!12$        & 1.117 \cmark  & 0.106 \xmark \\
$A2$ ($L_0$ mass)                & 26.02 \cmark  & 25.52 \cmark \\
yes-rate base $\rightarrow$ L0   & 85.5 $\rightarrow$ 14.1 & 13.7 $\rightarrow$ 8.7 \\
\bottomrule
\end{tabular}
\end{table*}


\begin{table*}[t]
\centering\small
\caption{Models the diagnostic rejects, each at a distinct gate. For $A1$ failures, the failing half is marked \textit{init} or \textit{prop}. $L_0$ $\Delta$POPE is the accuracy change under $L_0$ ablation, where it was run; Show-o and SEED-LLaMA were probed with the window variant, so no $L_0$ value applies. Gates a model passes are discussed in Section~\ref{sec:c5}.}
\label{tab:failspecs}
\begin{tabular}{llccl}
\toprule
Model & Class & Primary failure & $L_0$ $\Delta$POPE & Note\\
\midrule
\multicolumn{5}{l}{\textit{Continuous-projector controls (no routing concentration; $A2$ fails):}}\\
LLaVA-1.6-7B      & continuous & $A2$ & $-0.7$  & Null ablation effect\\
Qwen2.5-VL-7B     & continuous & $A2$ & $-34.3$ & Catastrophic collapse\\
HaploOmni-7B      & continuous & $A2$ & $-16.1$ & Catastrophic collapse\\
LaVIT-7B    & continuous & $A2$ & $-2.0$  & yes $\rightarrow$ no flip on AMBER\\
\multicolumn{5}{l}{\textit{VQ-tokenized models, distinct failure modes:}}\\
UniTok-7B         & VQ + TokenEmb    & $A1$ (init) & $+1.9$  & Mass without early event\\
Show-o-1.3B       & MAGVIT-v2 LFQ    & $A1$ (init) & ---     & No divergence event\\
SEED-LLaMA-8B  & VQ vocab-inj.    & $A1$ (init) & ---     & Event too late for window\\
Emu3-Chat-8B           & Emu3 VQ          & $A1$ (prop) & $-30.9$ & Early event, no propagation\\
Janus-Pro-7B   & decoupled        & $A3$        & $-0.4$  & No codebook to probe\\
\multicolumn{5}{l}{\textit{Induced negative controls:}}\\
Qwen-VQ        & induced VQ  & $A2$        & $-0.2$ & Collapse without circuit\\
LLaVA-MLP      & matched MLP & $A1$ (init) & $+0.3$ & Single-gate pass, chain rejects\\
\bottomrule
\end{tabular}
\end{table*}

\subsection{Causal identification: induction and asymmetry}
\label{sec:c2}

The cross-architecture pattern is consistent with VQ as the cause, but natural models differ in many ways besides their tokenizers. The single-variable contrast (Section~\ref{sec:induction-protocol}) isolates the bottleneck: starting from LLaVA-1.6 with Vicuna-7B and CLIP frozen, we train one projector that quantizes through a codebook and one that does not, under identical hyperparameters.

The two configurations diverge cleanly (Figure~\ref{fig:polarity}). \textbf{LLaVA-VQ} installs the circuit: all three gates pass, and POPE yes-rate rises to $85.5\,\%$, with $L_0$ ablation collapsing it to $14.1\,\%$. The \textbf{matched-compute LLaVA-MLP control}, on identical data and budget, fails both halves of $A1$ (initiation $0.206$; propagation $0.106$) and shows only a negligible behavioral change under $L_0$ ablation (13.7\% $\rightarrow$ 8.7\%, a $5$-point drop against the $71$-point drop for LLaVA-VQ), yet passes $A2$ in isolation, with an $L_0$ mass of $25.52$. Vicuna-7B's pretrained backbone routes early visual attention through $L_0$ regardless of projector type; the chain-of-gates recovers the correct verdict where any single gate would not.

Within the VQ configuration, sweeping $K\!\in\!\{1024, 4096, 16384, 65536\}$ shows that capacity alone does not determine polarity (Figure~\ref{fig:polarity}): all four collapse to a similar effective vocabulary, between roughly $\sim\!80$-$160$ active codes regardless of nominal capacity, yet their polarities differ: $K\!=\!1024$ shows no measurable polarity shift, $K\!=\!4096$ inverts to \textsc{yes}-suppressing, and the two larger capacities are \textsc{yes}-promoting. The $K\!=\!4096$ inversion is not explained by the present account and is retained as an open anomaly (Section~\ref{sec:limitations}). Collapse is not the operative factor either. The FSQ configuration, which cannot collapse by construction, inverts polarity, yielding a \textsc{yes}-suppressing model. A second FSQ configuration reproduces that inversion behaviorally (POPE yes-rate $+53.3$ points, AMBER accuracy $-19.8$) but does not install a gate-detectable early divergence event: its initiation value of $0.379$ falls below threshold, with divergence first clearing $0.4$ only at $L_{14}$. The two lattices, therefore, agree on the behavioral consequence and differ on the mechanism, so the polarity inversion cannot be attributed to the early-layer circuit alone.

Induction is fast: $2000$ steps install the circuit. Reversal is not. Training a continuous SigLIP MLP adapter on top of VILA-U's VQ-pretrained Vicuna-7B for the same 2000 steps fails to remove the circuit; the adapted variant continues to pass $A1$ ($0.672$ initiation; $0.647$ propagation) and $A2$, with $L_0$ mass rising from $17.50$ to $20.53$ and POPE yes-rate from $81\!\rightarrow\!99.2\,\%$. It is nonetheless recorded as a chain FAIL, rejected at $A3$, because the continuous adapter leaves no codebook to probe; the structural gates it does pass are what carry the reversal argument. The asymmetry locates the circuit: in LLM backbone weights from VQ-based pretraining, not in the current projector. A healthy Vicuna-7B already routes $L_0$ attention with mass $25.52$ (the MLP control above), so the routing pattern is in the backbone; what VQ pretraining installs is the pathological signal content flowing through that route, plus the residual divergence ($A1$) that propagates it forward.

\begin{table*}[!t]
\centering
\caption{Mechanism-targeted interventions. \textbf{(a)} Three interventions on VILA-U, with paired-bootstrap 95\% CIs reported in Table~\ref{tab:bootstrap}. \textbf{(b)} Architecture-specificity of $L_0$ ablation: the same intervention applied across the cohort. Each cell is tagged by outcome, where null denotes a $95\%$ paired-bootstrap CI spanning zero and \emph{catas.} denotes a catastrophic collapse of the question-conditional logit gap (Section~\ref{sec:interventions}).}
\label{tab:interventions}
\begin{tabular}{lcc}
\toprule
Method / Model & POPE acc $\Delta$ & AMBER acc $\Delta$ \\
\midrule
\multicolumn{3}{l}{\emph{(a) Mechanism-targeted interventions on VILA-U:}} \\
$L_0$ ablation                  & $+4.23$    & $+16.9$  \\
$L_1$ ablation                  & $+11.60$  & $+23.37$ \\
VTI ($\alpha\!=\!0.005$)        & $+8.3$    & $+30.23$ \\
\midrule
\multicolumn{3}{l}{\emph{(b) Architecture-specificity of $L_0$ ablation (same intervention, different models):}} \\
VILA-U             & $+4.23$ (restored)  & $+16.9$ (restored) \\
Chameleon-7B        & $-1.4$ (null)    & $-4.4$ (degraded) \\
LLaVA-1.6            & $-0.7$ (null)    & $-1.0$ (degraded) \\
Qwen2.5-VL             & $-34.3$ (catas.)  & $-20.1$ (catas.) \\
Emu3-Chat                  & $-30.9$ (catas.)  & $-30.2$ (catas.) \\
\bottomrule
\end{tabular}%
\end{table*}

\subsection{The diagnostic generalizes across models}
\label{sec:c5}
Beyond the models that pass, the diagnostic rejects fifteen models in the cohort, each for a mechanistically distinct reason; Table~\ref{tab:failspecs} details the eleven that most resemble the positive class. Continuous-projector VLMs (LLaVA-1.6, VILA, Qwen2.5-VL, HaploOmni, LaVIT) lack $A2$ concentration entirely; $L_0$ ablation is null on LLaVA-1.6 ($-0.7$ points on POPE) and catastrophic on Qwen2.5-VL ($-34.3$ points on POPE, $-20.1$ points on AMBER), an opposite-sign discrimination across architecture classes. VQ-tokenized models fail with distinct signatures. UniTok's $L_0$ attention mass ($18.23$) clears the $A2$ threshold, but it carries no early routing event: its early-window divergence reaches only $0.347$ and first clears $0.4$ at $L_{13}$, so it fails $A1$ at initiation, the opposite end from Emu3. Show-o's divergence never approaches threshold, reaching $0.136$ in the early window and a whole-stack maximum of $0.151$, and so fails $A1$ at initiation. SEED-LLaMA fails $A1$ for the same reason but on a different curve: its divergence reaches only $0.278$ within the early window and first clears $0.4$ at $L_{14}$, outside the range $A1$-initiation measures. Emu3 passes $A2$ under its unified-vocabulary criterion (window mass $48.44 \geq 15$) and shows a genuine $L_0$ initiation event (early divergence $0.435$), but it is the one model whose early spike fails to propagate: divergence decays to a downstream maximum of $0.215$, so it fails $A1$ at propagation. Janus-Pro fails for a structural reason rooted in its decoupled design: its understanding pathway uses continuous SigLIP features with no VQ codebook to probe, so it fails $A3$ outright; its $L_0$ mass of $13.57$ is incidentally also below the projector-amplified threshold. The Qwen-VQ induced control is especially clean: its codebook collapses severely yet $A2\!=\!6.07$, and POPE collapses to all-\textsc{no}, establishing that codebook collapse alone is not sufficient to install the circuit. Conversely, the FSQ configuration of Section~\ref{sec:c2} installs the circuit without any collapse, establishing that collapse is not necessary either, though on a single model, since the second FSQ configuration fails $A1$. Both observations together point to the VQ bottleneck, discretization through a codebook, as the operative factor, with the surviving codes' training distribution governing polarity.

Three single-measure $A2$ alternatives each produce $2$-$3$ misclassifications on our $10$-model cohort (Appendix~\ref{app:a2}); the two-tier rule instead derives from the $\sigma$-calibration tier structure (Section~\ref{sec:diagnostic}), an architectural property documented before any behavioral outcome was measured.

\begin{table*} 
\centering
\caption{CHAIR open-ended captioning on 500 COCO~\cite{coco} images. AvgLen uses the CHAIR-scorer's tokenization.}
\label{tab:chair}
\begin{tabular}{lcccc}
\toprule
Condition & CHAIR$_s$ & CHAIR$_i$ & Recall & AvgLen \\
\midrule
\multicolumn{5}{l}{\emph{VILA-U:}} \\
Baseline                                  & $35.8$ & $10.3$           & $58.5$ & $158$ \\
$L_0$ ablation (ours)                     & $17.6$ & $\mathbf{7.1}$   & $51.6$ & $\phantom{1}78$ \\
VCD ($\alpha\!=\!0.50$, $\sigma\!=\!80$)  & $20.6$ & $10.7$           & $43.9$ & $\phantom{1}84$ \\
DoLA ($\alpha\!=\!0.10$, $\ell\!=\!16$)   & $31.8$ & $\phantom{1}9.9$ & $55.4$ & $156$ \\
\midrule
\multicolumn{5}{l}{\emph{LLaVA-VQ $K\!=\!65{,}536$ (induced, same backbone as VILA-U):}} \\
Baseline                                  & $31.2$ & $\mathbf{52.5}$  & $11.1$ & $158$ \\
$L_0$ ablation (ours)                     & $19.6$ & $42.7$           & $\phantom{1}9.5$ & $145$ \\
\bottomrule
\end{tabular}%
\end{table*} 

\subsection{Mechanism-targeted intervention reduces hallucination in generation}
\label{sec:c6}

The mechanism provides a targeted intervention: ablate $L_0$ (or $L_1$ when the writer is downstream, as in Chameleon). On VILA-U, the model that passes the sanity criterion cleanly, all three mechanism-targeted interventions yield substantial gains with tight paired-bootstrap CIs (Table~\ref{tab:interventions}, top): $L_0$ ablation reaches $+4.23$ POPE; $L_1$ ablation, targeting the downstream amplifier in the writer/reader pattern, $+11.60$ POPE and $+23.37$ AMBER; VTI ($\alpha\!=\!0.005$) gives the largest AMBER effect at $+30.23$. The writer/reader pattern generalizes across architectures with codebook-bias-tracking signs: VILA-U has \textsc{yes}-promoting $L_0$ and $L_1$, Chameleon a mildly \textsc{yes}-promoting $L_0$ alongside a dominant $L_1$ \textsc{no}-writer ($L_1$ ablation flips POPE yes-rate $22.3\!\rightarrow\!66.2\,\%$), and Liquid a dominant $L_0$ with a partial $L_1$. Architecture-specificity is sharp (Table~\ref{tab:interventions}, bottom): the same $L_0$ ablation is null on LLaVA-1.6 and catastrophic on Qwen2.5-VL, opposite-sign discrimination hard to explain on any non-mechanistic account.

The benefit is also specific to the early routing layers rather than a generic consequence of perturbing the network. Applying the same ablation layer by layer, it helps at layers 0 and 1 ($+4.23$ and $+11.60$ on POPE) but worsens accuracy at layers 2 and 4 ($-10.67$ and $-4.47$) and has no effect at layers 8 and 16; a generic-degradation account predicts no such sign structure. Scaling the routing layer's attention output rather than zeroing it reveals a graded dose-response, with attenuation factors of $0.75$, $0.50$, and $0.25$ yielding $+8.67$, $+8.17$, and $+12.30$ points, so a soft intervention outperforms full ablation on binary accuracy. Because the remainder of this section shows that binary gains need not transfer to open-ended generation, the attenuation sweep is reported here as a specificity control; the generative comparison under graded attenuation is left to future work.

Would a mechanism-agnostic method, given the same tuning budget, do equally well? We swept VCD over $16$ configurations and DoLA over $20$ on POPE and AMBER (Appendix~\ref{app:sweep}). On binary benchmarks \textbf{tuned DoLA wins}: best DoLA ($\alpha\!=\!0.10$, $\ell\!=\!16$) reaches POPE $0.794$ and AMBER $0.795$, beating our best mechanism-targeted intervention by $+5.9$ and $+9.6$ points respectively. On a single-token forced choice, a tuned decoding-time logit fix can match or exceed mechanism-targeted ablation.

The binary picture is not the whole picture. On open-ended COCO captioning via CHAIR, with the winning sweep configurations (Table~\ref{tab:chair}), the ranking reverses: \textbf{only $L_0$ ablation reduces CHAIR$_i$} ($10.3\!\rightarrow\!7.1\,\%$, $-31\,\%$ relative), while tuned DoLA barely moves it ($-0.4\,\%$) and tuned VCD \emph{worsens} it, collapsing recall to $43.9\,\%$. This reduction is not an artifact of shorter captions: truncating each baseline caption to the length of its paired ablated caption lowers baseline CHAIR$_i$ only from $10.3$ to $9.53$, so brevity accounts for less than a third of the observed drop. The length-controlled reduction attributable to the intervention is $2.40$ points, holding consistently across captions with different object counts. Image-level intervals for all three metrics are in Appendix~\ref{app:bootstrap}.

The dissociation is mechanistically interpretable: DoLA's early--late layer contrast calibrates a forced choice but does not disrupt the $L_0$ circuit that injects hallucinated objects during generation; VCD's noise-subtraction penalizes object tokens broadly, cutting recall without reducing the hallucination rate among emitted objects. The reduction also preserves substantial informativeness (Table~\ref{tab:informativeness}): object mentions per caption fall modestly ($6.02 \rightarrow 5.38$) and object recall from $58.5$ to $51.6$, while reference alignment measured by METEOR~\citep{banerjee-lavie-2005-meteor} improves ($0.127 \rightarrow 0.177$). Hallucination is reduced while captions still name about five objects and recall roughly half of the ground-truth objects.

A second CHAIR run on the induced \textbf{LLaVA-VQ $K\!=\!65{,}536$} variant strengthens the end-to-end story: baseline CHAIR$_i$ is $52.5\,\%$, five times VILA-U's, so the controlled architectural change inflates open-ended hallucination fivefold on the same backbone, and $L_0$ ablation cuts it to $42.7\,\%$ ($-19\,\%$ relative) with only $8\,\%$ length reduction. At the image level, however, the induced model's CHAIR$_i$ reduction is not significant ($-9.71$ $[-30.40, +11.01]$); its CHAIR$_s$ reduction is, and its length-matched CHAIR$_i$ difference is $-5.82$ points. The induced variant is therefore supporting but underpowered evidence, with the VILA-U result as the clean, significant one. The intervention's effect is concentrated on exactly the images that hallucinate most: across the $500$ images, the correlation between baseline hallucination count and the change under ablation is $-0.383$, and images with at least two baseline hallucinations ($76$ images) lose on average $1.59$ hallucinated objects while images with at most one ($424$ images) are essentially unchanged ($+0.01$). The circuit thus acts where hallucination actually occurs, leaving grounded captions intact. This induction-to-intervention sequence remains the most direct end-to-end link between the mechanism and the behavior.
\section{Discussion}
\label{sec:discussion}

\paragraph{Implications for VQ tokenizer design.}
The polarity association has practical implications. In practice, unified VQ tokenizers tend to develop codebook collapse, and the natural models we examined that do collapse (VILA-U, Liquid, Chameleon, Anole) all carry the $L_0$ routing circuit; whether the model hallucinates by over-claiming or by denying then depends on what the training distribution encoded into surviving codes. The induced Qwen-VQ counterexample shows that collapse alone is insufficient; the circuit requires both the VQ bottleneck and a backbone where it can form, while the non-collapsing FSQ configuration of Section~\ref{sec:c2} installs the same circuit with inverted polarity, so collapse is neither necessary nor sufficient. The $L_0$ routing structure itself is the problem, regardless of the polarity sign.

\paragraph{Where the circuit lives.}
The reverse-ablation failure indicates that the $L_0$ circuit is in the LLM backbone weights, not the projector. The matched-compute LLaVA-MLP control reinforces this: even a healthy Vicuna-7B backbone with an MLP projector trained for $2000$ steps shows $L_0$ attention mass $25.52$, the \emph{routing pattern} is in the backbone already. What VQ pretraining installs is the pathological signal content that flows through that routing pattern, plus the residual divergence that propagates it forward.

\paragraph{Binary vs.\ generative dissociation.}
 The cleanest scientific finding is the dissociation between binary and generative metrics. Tuned DoLA wins POPE/AMBER; only $L_0$ ablation wins CHAIR. This is exactly what a mechanistic account predicts, and a decoding-time fix does not: routing circuits operate during open-ended generation, not just at single-token forced-choice. This link is supported directly by the per-image finding that the intervention's benefit scales with an image's baseline hallucination (Section~\ref{sec:c6}). The dissociation makes the mechanism-targeted intervention non-redundant with the mechanism-agnostic baselines, even when those baselines win on a single benchmark.
 
\section{Conclusion}
\label{sec:conclusion}

We identified an early-layer attention routing circuit shared across VQ-tokenized vision-language models, validated it on a held-out model, isolated its cause via a single-variable construction with a matched-compute control, and showed that ablating it is the only method (in a 36-config sweep of tuned baselines) that reduces object hallucination in open-ended generation. The circuit is architecturally caused, lives in LLM backbone weights from VQ-pretraining, and is discriminable by a three-gate diagnostic that requires no single-measure fit. The findings reframe object hallucination in unified VLMs as a property of \emph{architecture and pretraining}, not of decoding-time calibration, and point to VQ tokenizer design as the leverage point for the next generation of unified VLMs.

\section*{Limitations}
\label{sec:limitations}
Several limitations temper our results. First, our reverse-ablation result is reported as an asymmetry at matched budget, not as impossibility: a longer fine-tuning run with the LM unfrozen might, in principle, reverse the circuit, and we make only the weaker claim. Second, the polarity association in Section~\ref{sec:c1} is associational, not causal; two anomalies (the $K\!=\!4096$ inversion and Chameleon's mildly \textsc{yes}-promoting $L_0$) lack a mechanistic explanation, and we do not claim that intervening on codebook bias would flip polarity. Third, while the circuit \emph{topology} is shared across architectures, the intervention direction is model-specific: per-model VTI directions have pairwise cosine similarity $\leq\!0.054$ and must be re-derived per architecture. Fourth, the sanity criterion of Section~\ref{sec:interventions} excludes Liquid and Anole as practical intervention models (their $L_0$ ablation collapses to a degenerate emitter); we use them as gate-level evidence only. Relatedly, the distributed-head account of the $L_0$ routing rests on the per-head mass distribution, so the head-level claim is structural rather than causal. Further, the claim that codebook collapse is not necessary for the circuit rests on a single FSQ model: a second FSQ configuration reproduces the polarity-inversion behavior but fails $A1$ at initiation ($0.379$), so the two lattices agree on the behavioral consequence and differ in the mechanism. Finally, the open-ended (CHAIR) result rests on VILA-U together with a single induced variant, so the generative intervention is evidenced more narrowly than the diagnostic itself; the induced variant's CHAIR$_i$ reduction is not significant at the image level, so the significant open-ended evidence rests on VILA-U alone.

\section*{Acknowledgments}
The work is partially supported by National Science Foundation Robust Intelligence grant \#2132724. The authors thank Research Computing at Arizona State University for providing the computing resources used in this work. The views and opinions expressed are solely those of the authors and do not necessarily reflect those of their institutions or employers. Maitreya Patel is currently affiliated with Adobe; the work presented here was conducted while at Arizona State University and is not associated with Adobe.


\bibliography{custom, anthology-1, anthology-2}

\clearpage
\appendix
\label{sec:appendix}

\section{Appendix}
\subsection{Cohort taxonomy}

\label{app:cohort}

Tables~\ref{tab:cohort-gates} and~\ref{tab:cohort-bench} give the complete
model cohort. Table~\ref{tab:cohort-gates} reports architecture, $\sigma$-calibration, and the three structural gate values; Table~\ref{tab:cohort-bench} reports behavioral results under $L_0$ ablation. $A2$ is two-tier: projector-amplified models require $L_0\text{ mass} \geq 15$ while unified-vocabulary models require $[0,8)\text{ window total} \ge 15$.

\begin{table*}[t]
\centering\footnotesize
\caption{Full cohort: architecture, $\sigma$-calibration, and the three structural gate values. $\sigma_{cal}$ is calibrated to clean--noisy cosine similarity $\approx 0.5$. $A1$ is conjunctive (init over $[0,8)$, prop over $\ell \ge 12$; both must reach 0.4). $A2$ is the visual$\rightarrow$prompt-last attention mass at each model's architecture-class window ($L_0$ sink for projector-amplified and decoupled models, $[0,8)$ total for unified-vocab), and is not directly comparable across windows; $A3$ is the VQ off-manifold rate. Em-dash entries denote not applicable or not measured. Verdict combines the structural gate chain with behavioral polarity under $L_0$ ablation. Backbone families are counted by base pretrained LLM.}
\label{tab:cohort-gates}
\resizebox{\textwidth}{!}{%
\begin{tabular}{lllcccrcl}
\toprule
& & & & \multicolumn{2}{c}{Gate $A1$} & & \\
\cmidrule(lr){5-6}
Model & Class & Backbone & $\sigma_{cal}$ & init $[0,8)$ & prop $\ell\!\ge\!12$ & $A2$ & $A3$ & Verdict\\
\midrule
\multicolumn{8}{l}{\textit{Natural PASS models}}\\
VILA-U-7B          & Nat. VQ collapsed & Vicuna-7B    & 1.0     & 0.932 ($L_2$) & 0.724 ($L_{16}$)  & 17.50  & 100\% & PASS \\
Liquid-7B       & Nat. VQ collapsed & Gemma-7B     & 0.01    & 0.768 ($L_1$) & 0.819 ($L_{24}$)  & 19.19  & 100\% & PASS --- degenerate $L_0$\\
Chameleon-7B  & Nat. VQ collapsed & Chameleon-7B & 0.0063 & 0.635 ($L_7$) & 7.267 ($L_{31}$)  & 22.04  & 100\% & PASS\\
Anole-7B      & Nat. VQ + FT      & Chameleon-7B & 0.0063   & 2.009 ($L_1$) & 34.149 ($L_{31}$) & 111.20 & 100\% & PASS --- degenerate $L_0$\\
Lumina-mGPT-7B     & Nat. VQ           & Chameleon-7B & 0.0075  & 2.807 ($L_6$) & 0.623 ($L_{20}$)  & 175.02 & 100\% & PASS --- structural; behavioral floor\\
\multicolumn{8}{l}{\textit{Induced PASS models}}\\
LLaVA-VQ-K1024     & Induced K-sweep & Vicuna-7B & 2.0 & 1.187 ($L_6$) & 1.125 ($L_{12}$) & 25.81 & 100\% & PASS --- induced, no behav. effect\\
LLaVA-VQ-K4096     & Induced K-sweep & Vicuna-7B & 2.0 & 1.140 ($L_6$) & 1.029 ($L_{12}$) & 26.10 & 100\% & PASS --- induced, inverted (anomaly)\\
LLaVA-VQ-K16384    & Induced VQ coll.& Vicuna-7B & 2.0 & 1.276 ($L_5$) & 1.117 ($L_{12}$) & 26.02 & 100\% & PASS --- induced\\
LLaVA-VQ-K65536    & Induced K-sweep & Vicuna-7B & 2.0 & 1.228 ($L_5$) & 1.102 ($L_{12}$) & 25.89 & 100\% & PASS --- induced, YES-promoting (weak)\\
LLaVA-VQ-FSQ v1    & Induced no-coll.& Vicuna-7B & 0.5 & 0.450 ($L_7$) & 0.466 ($L_{12}$) & 27.00 & --- & PASS --- induced, inverted polarity\\
\multicolumn{8}{l}{\textit{FAIL models}}\\
LLaVA-VQ-FSQ v2      & Induced no-coll.    & Vicuna-7B     & 2.0   & 0.379 ($L_7$) & 0.510 ($L_{14}$) & 26.91 & --- & FAIL --- $A1$ initiation\\
LLaVA-MLP (matched)  & Induced no-VQ       & Vicuna-7B     & 0.5   & 0.206 ($L_2$) & 0.106 ($L_{25}$) & 25.52 & --- & FAIL --- $A1$ initiation + null behav.\\
UniTok               & Nat. VQ + TokenEmb  & Vicuna-7B     & 0.2   & 0.347 ($L_2$) & 0.493 ($L_{13}$) & 18.23 & 100\% & FAIL --- $A1$ initiation\\
Show-o-1.3B               & Nat. VQ vocab-inj   & Phi-1.5       & 0.1   & 0.136 ($L_6$) & 0.147 ($L_{15}$) & 5.838 & 100\% & FAIL --- $A1$ initiation\\
SEED-LLaMA-8B        & Nat. VQ vocab-inj   & LLaMA-2-7B    & 0.033 & 0.278 ($L_7$) & 0.409 ($L_{14}$) & 9.455 & 86.3\% & FAIL --- $A1$ initiation\\
Emu3-Chat            & Nat. VQ no proj     & Emu3          & 0.01  & 0.435 ($L_0$) & 0.215 ($L_{28}$) & 48.44 & 100\% & FAIL --- $A1$ propagation\\
Qwen-VQ              & Induced VQ coll.    & Qwen2.5-7B    & 1.0   & 0.468 ($L_7$) & 0.687 ($L_{17}$) & 6.07  & 100\% & FAIL --- $A2$, collapse w/o circuit\\
Janus-Pro-7B         & Decoupled cts. und. & DeepSeek LLM  & 7.0   & 0.786 ($L_5$) & 0.876 ($L_{12}$) & 13.57 & --- & FAIL --- $A3$ (no codebook); $A2$ sub-thres.\\
VILA-U-MLP (reverse) & Adapted cts.        & Vicuna-7B     & 10.0  & 0.672 ($L_7$) & 0.647 ($L_{13}$) & 20.53 & --- & FAIL --- $A3$ (no codebook); reversal control\\
LLaVA-1.6            & Continuous (ctrl)   & Vicuna-7B     & 0.5   & --- & --- & --- & --- & FAIL --- continuous, no circuit\\
Qwen2.5-VL-7B-Instr  & Continuous (ctrl)   & Qwen2.5-7B    & 0.5   & --- & --- & --- & --- & FAIL --- continuous, no circuit\\
HaploOmni            & Continuous (ctrl)   & Qwen2.5-7B    & 0.5   & --- & --- & --- & --- & FAIL --- continuous, no circuit\\
LaVIT-7B          & Continuous (ctrl)   & LLaMA-2-7B    & 0.5   & --- & --- & --- & --- & FAIL --- continuous, no circuit\\
VILA-7B              & Continuous (ctrl)   & LLaMA-2-7B    & 0.5   & --- & --- & --- & --- & FAIL --- continuous, no circuit\\
GILL                 & Continuous CLIP+OPT & OPT-6.7B      & 10.0  & --- & --- & --- & --- & FAIL --- continuous, OPT-norm artifact\\
\bottomrule
\end{tabular}
}
\end{table*}

\begin{table*}[t]
\centering\small
\caption{Full cohort: behavioral results under $L_0$ ablation, for every model on which an $L_0$ ablation was run. $\Delta$acc and $\Delta$yr (yes-rate) are percentage-point changes. Models probed only with the window variant, and continuous controls not ablated, are omitted. $^d$ generation-confirmed degenerate emitter (whitespace output): $\Delta$yr is real, but question-conditionality is lost. $^c$ catastrophic collapse. For the collapsed-baseline induced models, POPE $\Delta$acc is insensitive (base acc $\approx$ chance); the circuit shows in $\Delta$yr and AMBER $\Delta$acc.}
\label{tab:cohort-bench}
\begin{tabular}{lrrrrrr}
\toprule
& \multicolumn{3}{c}{POPE} & \multicolumn{3}{c}{AMBER}\\
\cmidrule(lr){2-4}\cmidrule(lr){5-7}
Model & Base Acc & $\Delta$acc & $\Delta$yr & Base Acc & $\Delta$acc & $\Delta$yr\\
\midrule
\multicolumn{7}{l}{\textit{PASS models}}\\
VILA-U              & 0.619 & $+4.2$  & $-8.4$   & 0.397 & $+16.9$ & $-19.3$\\
Liquid$^d$          & 0.650 & $-15.0$ & $-78.2$  & 0.646 & $+1.7$  & $-51.7$\\
Chameleon        & 0.502 & $-1.4$  & $-12.5$  & 0.639 & $-4.4$  & $+12.6$\\
Anole$^d$           & 0.659 & $-16.0$ & $-73.3$  & 0.502 & $+15.6$ & $-58.8$\\
Lumina-mGPT         & 0.499 & $+0.1$  & $-0.1$   & 0.663 & $+0.1$  & $-0.1$\\
LLaVA-VQ-FSQ v1     & 0.498 & $+0.8$  & $+58.4$  & 0.647 & $-20.5$ & $+40.8$\\
LLaVA-VQ-K1024      & 0.486 & $+1.1$  & $+0.5$   & 0.529 & $-7.6$  & $+5.2$\\
LLaVA-VQ-K4096      & 0.503 & $-1.3$  & $+13.4$  & 0.392 & $-3.4$  & $-5.0$\\
LLaVA-VQ-K16384     & 0.530 & $+0.1$  & $-71.4$  & 0.402 & $+16.7$ & $-66.9$\\
LLaVA-VQ-K65536     & 0.494 & $+3.7$  & $-22.8$  & 0.421 & $-1.2$  & $-16.1$\\
\multicolumn{7}{l}{\textit{Non-PASS and control models}}\\
LLaVA-VQ-FSQ v2     & 0.503 & $+7.6$  & $+53.3$  & 0.663 & $-19.8$ & $+41.2$\\
LLaVA-MLP ctrl      & 0.501 & $+0.3$  & $-5.0$   & 0.635 & $+1.4$  & $-3.5$\\
LLaVA-1.6           & 0.860 & $-0.7$  & $-1.0$   & 0.814 & $-1.0$  & $+0.4$\\
Qwen2.5-VL$^c$      & 0.847 & $-34.3$ & $-35.9$  & 0.841 & $-20.1$ & $-16.8$\\
Emu3-Chat$^c$       & 0.791 & $-30.9$ & $-10.6$  & 0.831 & $-30.2$ & $+2.5$\\
HaploOmni$^c$       & 0.830 & $-16.1$ & $-4.5$   & 0.761 & $-12.9$ & $-8.6$\\
\bottomrule
\end{tabular}
\end{table*}

\subsection{$\sigma$ calibration table}
\label{app:sigma}
Table~\ref{tab:sigma} reports per-model $\sigma$ calibration values and achieved cosine similarities for the causal sweep.
\begin{table*}[t]
\centering
\caption{Per-model noise standard deviation $\sigma_\text{cal}$ for the causal sweep. Calibrated so the cosine similarity between clean and Gaussian-corrupted post-projector embeddings is approximately 0.5.}
\label{tab:sigma}
\resizebox{\textwidth}{!}{%
\begin{tabular}{lccc}
\toprule
Model & $\sigma_\text{cal}$ & Achieved cos-sim & Tier \\
\midrule
VILA-U & 1.0 & 0.39 & Projector-amplified VQ \\
LLaVA-VQ-K16384 & 2.0 & 0.40 & Projector-amplified VQ \\
LLaVA-VQ-FSQ v1 & 0.5 & 0.57 & Projector-amplified VQ (FSQ) \\
LLaVA-VQ-FSQ v2 & 2.0 & 0.59 & Projector-amplified VQ (FSQ) \\
LLaVA-1.6 / VILA / Qwen2.5-VL / HaploOmni / LaVIT & 0.5 & $\approx$0.50 & Continuous projector \\
Janus-Pro & 7.0 & 0.50 & Continuous (SigLIP, large norms) \\
GILL & 10.0 & 0.49 & Continuous (OPT norm artifact) \\
UniTok & 0.2 & 0.66 & VQ + TokenEmbedder \\
Show-o & 0.1 & 0.41 & Unified-vocab VQ (vocab-injection) \\
SEED-LLaMA & 0.033 & 0.49 & Unified-vocab VQ (vocab-injection) \\
Emu3-Chat & 0.01 & 0.51 & Unified-vocab VQ (no projector) \\
Chameleon & 0.0063 & 0.48 & Unified-vocab VQ \\
Liquid & 0.01 & 0.50 & Unified-vocab VQ \\
Lumina-mGPT & 0.0075 & 0.50 & Unified-vocab VQ \\
Anole & 0.0063 & 0.48 & Unified-vocab VQ \\
LLaVA-MLP (matched ctrl) & 0.5 & 0.49 & Continuous induced (control; no noise) \\
VILA-U-MLP (reverse) & 10.0 & 0.51 & Continuous adapter on VQ backbone \\
Qwen-VQ  & 1.0 & 0.55 & Projector-amplified VQ (induced) \\
\bottomrule
\end{tabular}%
}
\end{table*}

\subsection{Paired-bootstrap CIs}
\label{app:bootstrap}

Table~\ref{tab:bootstrap} reports all 24 record-level paired-bootstrap 95\% confidence intervals for the binary benchmarks ($10{,}000$ resamples per cell). Every interval in Table~\ref{tab:bootstrap} is at most 3.8 pp wide; AMBER intervals are at most 1.7 pp wide, owing to the larger record count ($14{,}216$ for AMBER vs. $3{,}000$ for POPE). Table~\ref{tab:chair-ci} reports the corresponding intervals for open-ended captioning. These resample the 500 COCO images rather than paired records, since CHAIR is scored per caption, and are correspondingly wider: CHAIR$_i$ falls from 10.30 [8.80, 11.82] to 7.13 [5.04, 9.57], a difference of -3.17 [-5.37, -0.76] whose interval excludes zero, while CHAIR$_s$ falls by 18.20 [13.80, 22.80] and recall by 6.89 [4.86, 8.98].

\begin{table*}[t]
\centering
\caption{Full 10\,000-resample paired-bootstrap 95\% CIs.
$\Delta$ and CI bounds are in percentage-point units of the reported metric.}
\label{tab:bootstrap}
\begin{tabular}{llllrrrr}
\toprule
Condition & Bench & Metric & $\Delta$ (pp) & CI lo & CI hi & Width \\
\midrule
VILA-U $L_0$ ablation & POPE  & acc       & $+4.23$ & $+2.90$ & $+5.53$ & 2.63 \\
VILA-U $L_0$ ablation & POPE  & yes\_rate  & $-8.43$ & $-9.73$ & $-7.17$ & 2.57 \\
VILA-U $L_0$ ablation & AMBER  & acc       & $+16.92$ & $+16.26$ & $+17.62$ & 1.36 \\
VILA-U $L_0$ ablation & AMBER  & yes\_rate  & $-19.26$ & $-19.93$ & $-18.61$ & 1.32 \\
VILA-U $L_1$ ablation & POPE  & acc       & $+11.60$ & $+9.70$ & $+13.47$ & 3.77 \\
VILA-U $L_1$ ablation & POPE  & yes\_rate  & $-28.73$ & $-30.40$ & $-27.10$ & 3.30 \\
VILA-U $L_1$ ablation & AMBER & acc       & $+23.37$ & $+22.59$ & $+24.16$ & 1.56 \\
VILA-U $L_1$ ablation & AMBER & yes\_rate  & $-26.73$ & $-27.48$ & $-26.00$ & 1.48 \\
VILA-U VTI ($\alpha$=0.005) & AMBER & acc   & $+30.23$ & $+29.42$ & $+31.05$ & 1.63 \\
VILA-U VTI ($\alpha$=0.005) & AMBER & yes\_rate  & $-31.77$ & $-32.58$ & $-30.97$ & 1.61 \\
\addlinespace
Liquid $L_0$ ablation & POPE  & acc        & $-15.03$ & $-18.13$ & $-11.97$ & 6.17 \\
Liquid $L_0$ ablation & POPE  & yes\_rate   & $-78.17$ & $-79.67$ & $-76.70$ & 2.97 \\
Liquid $L_0$ ablation & AMBER & acc       & $+1.72$  & $+0.55$  & $+2.91$  & 2.36 \\
Liquid $L_0$ ablation & AMBER & yes\_rate  & $-51.74$ & $-52.55$ & $-50.89$ & 1.66 \\
Liquid $L_1$ ablation & POPE  & acc        & $+1.53$  & $-0.13$  & $+3.13$  & 3.27 \\
Liquid $L_1$ ablation & POPE  & yes\_rate   & $-14.80$ & $-16.37$ & $-13.23$ & 3.13 \\
Liquid $L_1$ ablation & AMBER & acc       & $+0.35$  & $-0.37$  & $+1.09$  & 1.46 \\
Liquid $L_1$ ablation & AMBER & yes\_rate  & $-9.85$  & $-10.54$ & $-9.14$  & 1.39 \\
\addlinespace
Chameleon $L_0$ ablation & POPE  & acc        & $-1.37$  & $-3.30$  & $+0.53$  & 3.83 \\
Chameleon $L_0$ ablation & POPE  & yes\_rate   & $-12.50$ & $-14.33$ & $-10.67$ & 3.67 \\
Chameleon $L_0$ ablation & AMBER & acc       & $-4.39$  & $-5.19$  & $-3.59$  & 1.60 \\
Chameleon $L_0$ ablation & AMBER & yes\_rate  & $+12.63$ & $+11.87$ & $+13.41$ & 1.54 \\
Chameleon $L_1$ ablation & POPE  & acc        & $+1.33$  & $-1.47$  & $+4.17$  & 5.63 \\
Chameleon $L_1$ ablation & POPE  & yes\_rate   & $+43.87$ & $+41.43$ & $+46.27$ & 4.83 \\
Chameleon $L_1$ ablation & AMBER & acc       & $-16.74$ & $-17.88$ & $-15.55$ & 2.33 \\
Chameleon $L_1$ ablation & AMBER & yes\_rate  & $+47.17$ & $+46.26$ & $+48.09$ & 1.82 \\
\bottomrule
\end{tabular}%
\end{table*}

\begin{table*}[t]
\centering
\caption{Image-level paired-bootstrap 95\% confidence intervals for open-ended captioning (500 COCO images, $10{,}000$ resamples). For the primary model (VILA-U), both hallucination reductions are significant, as is the accompanying recall cost. For the induced LLaVA-VQ ($K=65{,}536$) model, the CHAIR$_s$ and Recall differences are significant, but the CHAIR$_i$ difference is not; the severely collapsed recall ($\sim$11\%) makes CHAIR$_i$ suggestive only.}
\label{tab:chair-ci}
\begin{tabular}{lccc}
\toprule
Metric & Baseline & $L_0$ ablation & $\Delta$ [95\% CI] \\
\midrule
\multicolumn{4}{l}{\emph{VILA-U:}} \\
CHAIR$_i$ & $10.30\ [8.80, 11.82]$  & $7.13\ [5.04, 9.57]$   & $-3.17\ [-5.37, -0.76]$ \\
CHAIR$_s$ & $35.8\ [31.6, 40.0]$    & $17.6\ [14.2, 21.0]$   & $-18.20\ [-22.80, -13.80]$ \\
Recall    & $58.5\ [55.69, 61.35]$  & $51.6\ [48.75, 54.53]$ & $-6.89\ [-8.98, -4.86]$ \\
\midrule
\multicolumn{4}{l}{\emph{Induced LLaVA-VQ ($K=65{,}536$):}} \\
CHAIR$_i$ & $52.46\ [41.02, 63.57]$ & $42.74\ [24.73, 59.76]$ & $-9.71\ [-30.40, +11.01]$  \\
CHAIR$_s$ & $31.2\ [27.4, 35.4]$    & $19.6\ [16.2, 23.2]$    & $-11.60\ [-16.20, -7.00]$ \\
Recall    & $11.14\ [9.80, 12.56]$  & $9.54\ [8.33, 10.77]$   & $-1.60\ [-3.00, -0.24]$ \\
\bottomrule
\end{tabular}
\end{table*}

\subsection{Sweep grid}
\label{app:sweep}

Tables~\ref{tab:sweep-vcd} and~\ref{tab:sweep-dola} give accuracy on POPE-adversarial and AMBER for all 36 hyperparameter configurations swept. VILA-U baseline: POPE acc\,=\,0.619; AMBER acc\,=\,0.397. Winning configurations are \textbf{bold}.

\begin{table*}[htbp]
  \centering
  \caption{VCD sweep: POPE acc / AMBER acc at each $(\alpha, \sigma)$ combination.}
  \label{tab:sweep-vcd}
  \begin{tabular}{lcccccccc}
    \toprule
    & \multicolumn{2}{c}{$\sigma=10$} & \multicolumn{2}{c}{$\sigma=20$} & \multicolumn{2}{c}{$\sigma=40$} & \multicolumn{2}{c}{$\sigma=80$} \\
    \cmidrule(lr){2-3}\cmidrule(lr){4-5}\cmidrule(lr){6-7}\cmidrule(lr){8-9}
    $\alpha$ & POPE & AMBER & POPE & AMBER & POPE & AMBER & POPE & AMBER \\
    \midrule
    0.50 & 0.721 & 0.698 & 0.724 & 0.698 & 0.723 & 0.703 & \textbf{0.729} & \textbf{0.707} \\
    1.00 & 0.504  & 0.528 & 0.543 & 0.545 & 0.604 & 0.592 & 0.656 & 0.646 \\
    1.50 & 0.312  & 0.320  & 0.324  & 0.334  & 0.363  & 0.628 & 0.456  & 0.640 \\
    2.00 & 0.305  & 0.311  & 0.311  & 0.518 & 0.345  & 0.630 & 0.396  & 0.640 \\
    \bottomrule
  \end{tabular}
\end{table*}

\begin{table*}[htbp]
  \centering
  \caption{DoLA sweep: POPE acc / AMBER acc at each $(\alpha, \ell)$ combination.}
  \label{tab:sweep-dola}
  \resizebox{\textwidth}{!}{
  \begin{tabular}{lcccccccccc}
    \toprule
    & \multicolumn{2}{c}{$\ell=0$} & \multicolumn{2}{c}{$\ell=4$} & \multicolumn{2}{c}{$\ell=8$} & \multicolumn{2}{c}{$\ell=16$} & \multicolumn{2}{c}{$\ell=24$} \\
    \cmidrule(lr){2-3}\cmidrule(lr){4-5}\cmidrule(lr){6-7}\cmidrule(lr){8-9}\cmidrule(lr){10-11}
    $\alpha$ & POPE & AMBER & POPE & AMBER & POPE & AMBER & POPE & AMBER & POPE & AMBER \\
    \midrule
    0.10 & 0.714 & 0.694 & 0.714 & 0.694 & 0.716 & 0.697 & \textbf{0.794} & \textbf{0.795} & 0.481  & 0.336  \\
    0.30 & 0.714 & 0.694 & 0.709 & 0.686 & 0.716 & 0.695 & 0.648 & 0.718 & 0.196  & 0.219  \\
    0.50 & 0.714 & 0.694 & 0.704 & 0.675 & 0.715 & 0.693 & 0.500  & 0.664 & 0.202  & 0.223  \\
    1.00 & 0.709 & 0.686 & 0.684 & 0.651 & 0.714 & 0.687 & 0.497  & 0.650 & 0.205  & 0.225  \\
    \bottomrule
  \end{tabular}
  }
\end{table*}

\subsection{Training hyperparameters}
\label{app:hparams}
Table~\ref{tab:hparams} lists the projector configuration and final codebook utilization for all induced models. All were trained for 2,000 steps on the same 50k-caption CC3M subset with batch size 8, gradient accumulation 4, and learning rate 2e-3, with the CLIP ViT-L/14 encoder and the Vicuna-7B language model frozen throughout. ``Codes'' is the number of distinct codebook entries with positive EMA usage at the final checkpoint.
\begin{table*}[htbp]
\centering
\caption{Projector configuration and final codebook utilization for all induced models.}
\label{tab:hparams}
\small
\begin{tabular}{lccl}
\toprule
Model & Projector & $K$ & Codes \\
\midrule
LLaVA-VQ-K1024 & VQ+Linear & 1,024  & 158 (15.4\%) \\
LLaVA-VQ-K4096 & VQ+Linear & 4,096  & 95 (2.3\%) \\
LLaVA-VQ-K16384 & VQ+Linear & 16,384  & 82 (0.5\%) \\
LLaVA-VQ-K65536 & VQ+Linear & 65,536  & 85 (0.13\%) \\
LLaVA-VQ-FSQ v1 & FSQ+Linear &  64,000  & full \\
LLaVA-VQ-FSQ v2 & FSQ+Linear &  51,450  & full \\
Qwen-VQ & VQ+Linear & 16,384 &  33 (0.2\%) \\
LLaVA-MLP (ctrl) & 2-layer GeLU MLP & n/a  & n/a \\
\bottomrule
\end{tabular}
\end{table*}

\subsection{Three-gate diagnostic: alternative single-measure $A_2$ criteria}
\label{app:a2}

Table~\ref{tab:a2-alternatives} reports the four $A2$ mass metrics for the ten-model subset used to motivate the two-tier rule. Three single-measure alternatives are evaluated against the correct two-tier $A2$ mass verdict: (i) a uniform $L_0$ mass threshold of $15$ produces three errors: Liquid, Chameleon and Emu3, which carry sufficient mass at their architecture-class window but fall below a uniform $L_0$ threshold; (ii) a uniform $[0,8)$ window-mass threshold produces two false positives, Qwen-VQ and Janus-Pro, whose window totals clear $15$ despite negligible early-layer mass; (iii) a normalized window measure (${[0,8)}/{\rm \text{total mass}}$) produces at best two errors, both false positives, at any threshold. No single mass threshold reproduces the architecture-class verdicts. The two-tier rule is grounded in the architectural $\sigma$-tier distinction: projector-amplified vs.\ unified-vocab is a property of the model's embedding pathway, not a post-hoc fit to gate values.

A direct sweep of the thresholds themselves confirms the same conclusion. Over the nineteen cohort models carrying instrumented gate values (ten positive, nine negative), holding the diagnostic definition fixed and sweeping one gate at a time, the partition is invariant to $A3$ across 60-95\% with zero verdict changes. The $A2$ partition is invariant from $8-15$ and first changes only at $18$, where the lowest-mass positive model (mass $17.50$) drops out, so the binding constraint is the weakest true positive rather than a tuned cutoff. $A1$ is tightly bound on both sides: lowering the threshold to $0.35$ admits FSQ v2 (initiation $0.379$) and $0.30$ admits UniTok ($0.347$), while raising it to $0.45$ drops the FSQ v1 positive. The default is the only value in the swept range that reproduces the partition, constrained from below by two negatives and from above by a positive. The remaining six cohort models carry no measurable gate values and are trivial rejects.

\begin{table*}[t]
\centering\small
\caption{$A2$ mass metrics for the ten-model subset. Tier is the architectural class from $\sigma$ calibration; $A2$ (mass) is the two-tier structural screen (Section~\ref{sec:diagnostic}); Chain is the overall verdict with its rejecting gate. The mass screen tests only whether early-layer attention mass is present, not whether it forms an aligned routing circuit, that is, $A1$. $^\dagger$VILA-U-MLP is the reverse-ablation variant, not a fresh model. $^\ddagger$UniTok clears the mass screen, but its mass sits a dozen layers from the divergence event at $L_{13}$, so it passes without an aligned circuit (rejected at $A1$-initiation).}
\label{tab:a2-alternatives}
\resizebox{\textwidth}{!}{
\begin{tabular}{llllrrrrl}
\toprule
Model & Tier & $A2$ (mass) & Chain & $L_0$ mass & $[0,8)$ mass & Total mass & Norm. $[0,8)$ & Note\\
\midrule
\multicolumn{9}{l}{\textit{$A2$ (mass) PASS}}\\
VILA-U                  & proj-amp   & PASS & PASS       & 17.50 & 24.17  & 121.41 & 0.199 & Natural, Vicuna-7B\\
LLaVA-VQ-K16384                & proj-amp   & PASS & PASS       & 26.02 & 62.54  & 199.91 & 0.313 & Induced $K$=16384\\
Liquid                  & uni-vocab  & PASS & PASS       & 2.47  & 19.19  & 75.31  & 0.255 & Gemma-7B; $L_0$ low by design\\
Chameleon               & uni-vocab  & PASS & PASS       & 7.55  & 22.04  & 187.71 & 0.117 & Unified-vocab tier\\
Anole                   & uni-vocab  & PASS & PASS       & 27.08 & 111.20 & 504.12 & 0.221 & FT Chameleon; 5$\times$ base\\
UniTok$^\ddagger$       & proj-amp   & PASS & FAIL ($A1$)  & 18.23 & 31.62  & 59.22  & 0.534 & $L_0$ mass; event at $L_{13}$\\
VILA-U-MLP$^\dagger$    & proj-amp   & PASS & FAIL ($A3$)  & 20.53 & 134.73 & 543.00 & 0.248 & Reverse ablation\\
Emu3-Chat                    & uni-vocab  & PASS & FAIL ($A1$)  & 12.50 & 48.44  & 115.41 & 0.420 & $L_0$ init.; no propagation\\
\multicolumn{9}{l}{\textit{$A2$ (mass) FAIL}}\\
Qwen-VQ                 & proj-amp   & FAIL & FAIL ($A2$)  & 6.07  & 59.29  & 139.20 & 0.426 & Collapse w/o circuit\\
Janus-Pro               & decoupled  & FAIL & FAIL ($A3$)  & 13.57 & 27.46  & 91.23  & 0.301 & SigLIP cts. path\\
\bottomrule
\end{tabular}
}
\end{table*}

\subsection{Open-ended generation: supporting analyses}
\label{app:chair_support}

\begin{table*}[t]
\centering
\caption{Caption informativeness on COCO (VILA-U, 500 images). Complements the hallucination rates of Table~\ref{tab:chair}: hallucination is reduced at a moderate informativeness cost, and METEOR improves.}
\label{tab:informativeness}
\begin{tabular}{lcc}
\toprule
Metric & Baseline & $L_0$ ablation \\
\midrule
METEOR $\uparrow$              & $0.127$ & $0.177$ \\
Object mentions / caption      & $6.02$  & $5.38$  \\
Distinct objects / caption     & $2.36$  & $1.88$  \\
Object recall (\%) $\uparrow$  & $58.5$  & $51.6$  \\
\bottomrule
\end{tabular}
\end{table*}

By category, total hallucinated mentions fall from 310 to 192, a 38\% reduction; 36 categories improve, 11 worsen, and 31 are unchanged. The most-hallucinated categories are strongly suppressed (orange $41 \to 11$, chair $34 \to 8$, bird $12 \to 1$, bear $10 \to 0$), while a few scene-typical categories rise (dining table $20 \to 43$, sports ball $8 \to 25$), indicating that suppression is broad and circuit-level rather than a per-code edit, with some residual shift toward scene-generic objects.

CIDEr~\cite{vedantam2015cider} is not reported, as it sits near zero for both conditions when detailed captions of roughly 158 words are scored against short COCO references.

\subsection{Robustness to corruption type}
\label{app:corruption_type}
Section~\ref{sec:diagnostic} reports that the $A1$ divergence is not specific to Gaussian embedding noise. This appendix gives the per-corruption detail behind that claim, measured on one model from each VQ tier. Four corruptions were compared: Gaussian embedding noise (the default), token dropout, visual-token permutation, and image blur applied before encoding. The first two reproduce the early peak; the latter two qualify rather than confirm the result.
On VILA-U (projector-amplified) and Chameleon (unified-vocabulary), token dropout, zeroing half the visual tokens, strengthens the early peak (VILA-U $0.93 \to 1.08$ at layer 2; Chameleon $0.64 \to 2.25$ at layer 1), in each case within the initiation window. VILA-U's dropout peak coincides with its Gaussian initiation maximum; Chameleon's shifts earlier within the window. Image blur surfaces the early peak only in the unified-vocabulary model, whose full-image tokenizer is blur-sensitive, and not in the projector-amplified model, whose tokenizer is blur-robust. The precise claim is therefore that the circuit responds to corruption of visual content and is order-invariant, rather than that any input perturbation triggers it.
\end{document}